\documentclass{article}

\PassOptionsToPackage{numbers,sort&compress}{natbib}
\usepackage[eandd,preprint]{neurips_2026}

\usepackage{amsmath,amssymb,amsfonts}

\usepackage{booktabs}
\usepackage{array}
\usepackage{multirow}
\usepackage{makecell}

\usepackage{graphicx}
\usepackage[dvipsnames]{xcolor}

\usepackage{pifont}

\usepackage{enumitem}

\usepackage{algorithm}
\usepackage{algorithmic}

\usepackage{hyperref}
\hypersetup{hidelinks}
\usepackage{url}

\usepackage{subcaption}

\usepackage{tikz}
\usepackage{pgfplots}
\pgfplotsset{compat=1.17}
\usetikzlibrary{positioning, arrows.meta, shapes.geometric, fit, calc, backgrounds}
\usepgfplotslibrary{fillbetween}
\usepgfplotslibrary{groupplots}

\newcommand{\cmark}{\ding{51}}
\newcommand{\xmark}{\ding{55}}
\newcommand{\pmark}{\ding{73}}

\newcommand{\carlaair}{\textsc{CARLA-Air}}

\title{Can Aerial VLA Models Cooperate? Evaluating Closed-Loop Air-Ground Coordination with \carlaair{}}

\author{
Tianle Zeng\textsuperscript{1} \quad Yanci Wen\textsuperscript{1} \quad Xueang Yu\textsuperscript{2} \quad Hong Zhang\textsuperscript{1}\thanks{Corresponding author: Hong Zhang. Contact email: \texttt{12531293@mail.sustech.edu.cn}.} \\ \textsuperscript{1}Southern University of Science and Technology \\ \textsuperscript{2}Fudan University
}

\begin{document}

\maketitle

\begin{abstract}
Recent aerial vision-language-action (VLA) models show promising
single-UAV capabilities, such as tracking moving objects and navigating to
language-specified landmarks. However, it remains unclear whether these
capabilities can transfer to air-ground cooperation, where a UAV and a UGV
must act jointly in a shared, closed-loop physical world.

We study this question with \carlaair{}, a single-process air-ground
evaluation environment that unifies CARLA and AirSim inside one Unreal
Engine runtime. By sharing the same world state, physics tick, and sensing
pipeline, \carlaair{} enables physically consistent UAV--UGV interaction
and precise measurement of simulation-timestamp alignment and effective
coordination latency.

Using \carlaair{}, we evaluate representative aerial VLA and planning
baselines on two complementary diagnostic tasks: moving-platform landing
and occlusion-recovery escort. The results show that current aerial VLA
models can often track or follow a ground partner, but struggle to convert
this single-agent competence into stable cooperative behavior. State
prompting provides limited benefit, and naive bidirectional interaction
fails to consistently improve performance and can amplify errors for most baselines. 
These findings suggest that, under the tested text-based cue interfaces, zero-shot cooperative air-ground VLA requires three components beyond the current paradigm: explicit partner-state grounding, low-latency action coordination, and team-level objective alignment.
Our code is available at \url{https://github.com/louiszengCN/CarlaAir}
\end{abstract}.

\section{Introduction}
\label{sec:intro}

Air-ground cooperation is becoming an important capability for embodied
intelligence in urban search-and-rescue, autonomous logistics, inspection,
and intelligent transportation~\cite{chai2024cooperative}. A UAV can
provide wide-area aerial perception, while a UGV can execute ground-level
actions and carry heavier payloads~\cite{zeng2025ezreal}. In principle, the two platforms are
highly complementary. In practice, however, turning this complementarity
into real cooperation requires agents to share state, coordinate actions,
and respond to each other in a closed-loop physical world~\cite{chen2026vision}.

Recent aerial vision-language-action (VLA) models provide a tempting
starting point. They can follow language instructions, track moving ground
objects, and navigate toward visual goals from onboard observations
~\cite{uav_track_vla_2025,gao2025openfly,sautenkov2025uav,liu2023aerialvln}.
These abilities resemble useful components of air-ground cooperation.
This raises a central question: can single-UAV VLA competence naturally
transfer to cooperative UAV--UGV behavior?

We argue that this question remains largely unanswered. Existing
air-ground datasets~\cite{uav3d} mainly evaluate offline perception, while existing
interactive simulators~\cite{transimhub2025,airsimag2026} often rely on bridge-based or multi-process
architectures with independent clocks. 
Such settings make it difficult to measure whether one agent's signal actually improves the partner's real-time behavior, or whether apparent gains or failures are partly caused by clock mismatch and communication delay introduced by the evaluation platform itself.
A reliable evaluation of cooperative VLA
requires a shared physical world, closed-loop interaction, and measurable
end-to-end coordination latency.

To this end, we introduce \carlaair{}~\cite{zeng2026carla}, a single-process air-ground
evaluation environment that unifies CARLA and AirSim inside one Unreal
Engine runtime. \carlaair{} preserves CARLA's urban traffic simulation and AirSim's multirotor dynamics while placing UAVs and UGVs under the same world state, physics tick, and sensing pipeline. 
This design enables physically consistent interaction and removes the simulation-timestamp mismatch that can arise when UAV and UGV simulators are bridged as separate processes.

Built on \carlaair{}, we evaluate whether existing aerial VLA models can
move beyond single-agent behavior toward air-ground cooperation. We design two complementary diagnostic tasks. Cooperative Moving-Platform Landing tests whether visible tracking of a ground partner can become coordinated action. Cooperative Occlusion-Recovery Escort tests whether partner-state cues can help recover visual contact after temporary occlusion. Together, the two tasks probe cooperation from both the action side and the perception side, while keeping the native aerial VLA interface unchanged.

Our results reveal a consistent gap between aerial competence and
cooperation. Current aerial VLA models can often track or follow the UGV,
but this ability does not reliably convert into successful cooperative
behavior. Partner-state prompting brings limited and unstable gains, and
naive bidirectional interaction can even amplify errors. In contrast, a
state-based cooperative reference performs substantially better when
explicit metric state and low-latency coordination are available. 
These findings suggest that single-UAV VLA competence is not cooperation:
cooperative air-ground VLA requires mechanisms for partner-state modeling, cooperative observation design, partner-aware action grounding, and team-level objective alignment.

\textbf{Contributions.}
\begin{enumerate}[leftmargin=*, itemsep=2pt]
  \item \textbf{Closed-loop evaluation runtime.}
  We introduce and validate \carlaair{}, a single-process air-ground
  evaluation runtime that unifies CARLA and AirSim for closed-loop UAV--UGV
  interaction. By sharing world state, physics ticks, and sensing pipelines,
  \carlaair{} enables physically consistent interaction with zero
  simulation-timestamp mismatch by construction. This provides a reliable
  basis for closed-loop evaluation of air-ground cooperation without
  confounding the results with simulator synchronization errors.

  \item \textbf{Diagnostic evaluation suite.}
  We introduce two complementary diagnostic tasks, moving-platform landing and occlusion-recovery escort, together with communication protocols, baselines, a state-based cooperative reference, and metrics that separate single-UAV competence from cooperative success. Together, these components form a focused diagnostic suite for isolating cooperation failure modes rather than ranking general-purpose autonomy systems.

  \item \textbf{Scientific finding.}
  Under zero-shot evaluation and tested text-based cue protocols, current
  aerial VLA models do not naturally transfer single-UAV competence to
  air-ground cooperation: tracking does not reliably become coordination,
  partner cues do not consistently become recovery behavior, and naive
  interaction can amplify errors for most baselines.
\end{enumerate}

\section{Related Work}
\label{sec:related}
\begin{table}[t]
\centering
\caption{%
\textbf{Representative comparison of air-ground evaluation infrastructure.}
Column criteria: \emph{Single-proc.\ physics} = UAV and UGV share
one simulation clock and physics tick; \emph{Closed-loop coop.} = agents
interact in a live closed loop; \emph{Urban traffic} = dynamic ground
traffic; \emph{VLA eval.} = native support for vision-language-action
policy evaluation; \emph{Latency meas.} = measurable end-to-end
cooperation latency.
\carlaair{} is the only interactive platform combining all five properties.
}
\label{tab:competitor}
\small
\setlength{\tabcolsep}{4pt}
\begin{tabular}{lccccc}
\toprule
\textbf{System} &
  \makecell{\textbf{Single-proc.}\\\textbf{physics}} &
  \makecell{\textbf{Closed-loop}\\\textbf{coop.}} &
  \makecell{\textbf{Urban}\\\textbf{traffic}} &
  \makecell{\textbf{VLA}\\\textbf{eval.}} &
  \makecell{\textbf{Latency}\\\textbf{meas.}} \\
\midrule
\multicolumn{6}{@{}l}{\emph{Interactive simulation platforms}} \\
\carlaair{} (ours)                             & \cmark & \cmark & \cmark & \cmark & \cmark \\
Bridge (ROS arch.)                             & \xmark & \pmark & \pmark & \xmark & \xmark \\
TranSimHub~\cite{transimhub2025}               & \xmark & \cmark & \cmark & \xmark & \xmark \\
AirSimAG~\cite{airsimag2026}                   & \cmark & \pmark & \pmark & \xmark & \xmark \\
SimWorld-Robotics~\cite{zhuang2025simworld}    & \xmark & \cmark & \cmark & \cmark & \xmark \\
UnrealZoo~\cite{zhong2025unrealzoo}            & \xmark & \xmark & \pmark & \xmark & \xmark \\
OmniDrones~\cite{xu2024omnidrones}             & \cmark & \xmark & \xmark & \xmark & \xmark \\
gym-pybullet-drones~\cite{panerati2021learning}& \cmark & \xmark & \xmark & \xmark & \xmark \\
RotorS~\cite{furrer2016rotors}                 & \xmark & \xmark & \xmark & \xmark & \xmark \\
\midrule
\multicolumn{6}{@{}l}{\emph{Dataset / offline evaluation works (included for context)}} \\
UAV3D~\cite{uav3d}            & \xmark & \xmark & \cmark & \xmark & \xmark \\
Griffin~\cite{griffin2026}     & \xmark & \xmark & \cmark & \xmark & \xmark \\
OpenFly~\cite{gao2025openfly}    & N/A    & \xmark & \xmark & \cmark & \xmark \\
UAV-Flow~\cite{wang2025uavflow}  & N/A    & \xmark & \xmark & \cmark & \xmark \\
\bottomrule
\multicolumn{6}{l}{%
  \cmark~=~Yes;\quad \xmark~=~No;\quad \pmark~=~Partial;\quad
  N/A~=~not applicable.}
  \vspace{-0.55cm}
\end{tabular}
\end{table}

\subsection{Air-Ground Evaluation Infrastructure}
\label{ssec:rel_infra}
Existing evaluation infrastructure cannot provide reliable end-to-end
assessment of closed-loop cooperative behavior. Current efforts fall into two categories, each with fundamental
limitations.

\textbf{Air-ground cooperation benchmarks.}
Several air-ground datasets and benchmarks study cooperation between
aerial and ground agents, including UAV3D~\cite{uav3d},
CoPerception-UAVs~\cite{coperception_uav},
AirV2X~\cite{airv2x2025}, and Griffin~\cite{griffin2026}---but all
evaluate perception modules (detection, tracking, feature fusion) on pre-collected datasets.
Because agents do not interact in a live closed loop, these benchmarks cannot reveal whether one agent's signal improves the partner's real-time behavior or changes the final cooperative outcome.

\textbf{Air-ground simulation platforms.}
Existing interactive platforms either rely on multi-process or bridge-based architectures with independent clocks, or lack the full combination of dynamic urban traffic, VLA evaluation, and latency-measurable closed-loop cooperation.
The standard approach couples CARLA~\cite{carla2017} and
AirSim~\cite{airsim2017} or RotorS~\cite{furrer2016rotors} via a ROS bridge, introducing timing jitter
that corrupts latency measurement.
TranSimHub~\cite{transimhub2025} supports complex air-ground traffic
scenarios but retains the same multi-process synchronisation limitation.
Recent urban embodied AI platforms like SimWorld-Robotics~\cite{zhuang2025simworld} and UnrealZoo~\cite{zhong2025unrealzoo} provide rich, photo-realistic environments supporting both UAVs and ground vehicles, yet they lack the strict high-frequency physics synchronization required for precise air-ground control.
AirSimAG~\cite{airsimag2026} achieves single-process execution and supports basic UAV-UGV coordination, but lacks dynamic urban ground traffic and native support for evaluating VLA policies.
Similarly, reinforcement learning-oriented UAV platforms such as OmniDrones~\cite{xu2024omnidrones} and gym-pybullet-drones~\cite{panerati2021learning} offer scalable multi-agent training for aerial tasks, but are not designed for urban traffic realism or closed-loop ground interaction.
None provides the combination of a unified physical world,
closed-loop interaction, and precise latency instrumentation needed to
reliably evaluate air-ground cooperation. \carlaair{} is designed to fill
this gap.

\subsection{Methods for Air-Ground Cooperation}
\label{ssec:rel_methods}

\textbf{Modular and task-specific approaches.}
Air-ground cooperation has been explored through modular and task-specific systems. \cite{cladera2025air} combine an LLM planner with metric navigation for language-specified aerial-ground missions, while cooperative perception methods such as Where2Comm~\cite{where2comm} learn communication-efficient feature sharing across agents.
Bidirectional trajectory optimization~\cite{bidiadaptive2026} couples UAV and moving-platform trajectories for agile landing, and GLIDE~\cite{glide2025} uses compact georeferenced messages for role-separated UAV-UGV teaming in search-and-rescue.
Learning-based approaches~\cite{commawareMARL2026} further train partner-aware policies through structured observation interfaces.
These works show that air-ground cooperation can be achieved
with task-specific protocols, state representations, and coordination
mechanisms, but their reliance on designed interfaces or task-specific
coordination pipelines limits generalization to open-ended language
instructions. 
This motivates studying end-to-end aerial VLA models as more general language-conditioned UAV-side agents for air-ground cooperation.

\textbf{End-to-end aerial VLA models.}
End-to-end vision-language-action models that unify perception, reasoning, and control have made rapid progress on the aerial side.
Methods~\cite{gao2025openfly,liu2023aerialvln} enable UAVs to navigate to ground-level landmarks from language instructions;
UAV-Track VLA~\cite{uav_track_vla_2025} achieves autonomous aerial
tracking of moving ground vehicles; and AerialVLA~\cite{aerialvla2025} extends language-conditioned control to complex flight manoeuvres involving ground-scene understanding.
Recent benchmarks like UAV-Flow~\cite{wang2025uavflow} further demonstrate that aerial VLAs can perform language-guided fine-grained trajectory control in real-world flights.
These models already perform tasks that closely resemble components of
air-ground cooperation, making them promising candidates as the UAV-side
agent in a cooperative system.
However, whether their single-platform competence transfers to multi-platform cooperation has not been evaluated. \carlaair{} provides both the platform and the diagnostic suite to conduct this evaluation.

\section{\carlaair{} Platform}
\label{sec:platform}
To support shared-world interaction and latency-measurable closed-loop
evaluation, \carlaair{} implements a single-runtime design in which aerial and ground agents share the same world state, simulation clock, and physics tick. 
The key design choice is to embed CARLA and AirSim into one engine-level runtime rather than synchronizing two simulator processes after execution. This section describes how the runtime resolves simulator
ownership, preserves native interfaces, and validates that the resulting
measurements are free from inter-process synchronization noise.

\subsection{Causally Consistent Runtime Architecture}
\label{sec:platform:arch}
\textbf{Resolving the single-authority constraint.}
The main obstacle is not simply co-locating two large codebases, but
reconciling two independent notions of simulation authority within one
runtime world. In Unreal Engine, each simulation world has a single
authoritative lifecycle controller, implemented as \texttt{GameMode}~\cite{sobchyshak2025pushing}.
Since CARLA and AirSim were both designed as standalone backends, a naive
merge would create two competing controllers over the same world state.
This is why conventional integrations place the two simulators in separate processes and synchronize them externally.

\carlaair{} resolves this conflict by exploiting an asymmetry in how the
two backends depend on runtime authority. CARLA behaves as a
\emph{controller-dominant} backend: its traffic manager, weather system,
episode management, sensor scheduling, and RPC services are tightly bound
to the global lifecycle controller. AirSim, in contrast, is
\emph{actor-realizable}: its multirotor dynamics and flight control logic
are encapsulated in an actor-level vehicle entity and do not require
ownership of the global controller.

This asymmetry yields a single-authority, multi-actor runtime.
\carlaair{} preserves CARLA as the authoritative world manager and
instantiates AirSim as an actor-level aerial subsystem in the same
simulation world. Concretely, \texttt{CARLAAirGameMode} inherits CARLA's
native GameMode to retain ground-vehicle simulation, traffic, weather,
episode, sensor, and RPC functionalities, while composing the AirSim flight
actor as a standard world entity during engine initialization. Thus, the
integration changes the runtime ownership model without reimplementing
either simulator.

\textbf{Shared-clock execution.}
Timestamp alignment follows from the runtime structure rather than from an
additional synchronization module. Once the AirSim vehicle is composed as a
world actor, both the UGV and UAV are advanced by the same engine
scheduler. At each tick, the runtime applies ground and aerial commands,
advances the shared physics state, and samples sensors from the updated
world state. UAV and UGV observations are therefore generated from the same
state transition, not aligned after generation. This makes matched sensor
timestamps identical by construction in the single-threaded execution path.

\textbf{Native interfaces under a unified runtime.}
\carlaair{} changes the runtime boundary, not the user-facing API
boundary. CARLA and AirSim retain their original RPC servers and client
protocols, so existing ground and aerial policies can still issue commands
through the standard CARLA and AirSim Python APIs. The difference is that
both command streams are resolved inside the same runtime world: CARLA
commands act on the authoritative ground subsystem, while AirSim commands
act on the composed aerial actor. Native API compatibility is therefore
preserved without reintroducing an external bridge or independent
simulation clocks.

Coordinate-frame unification, sensor scheduling, synchronous rendering,
API compatibility, and hardware requirements are detailed in
Appendix~\ref{app:platform}.

\begin{figure*}[t]
    \centering
    \includegraphics[width=1\textwidth]{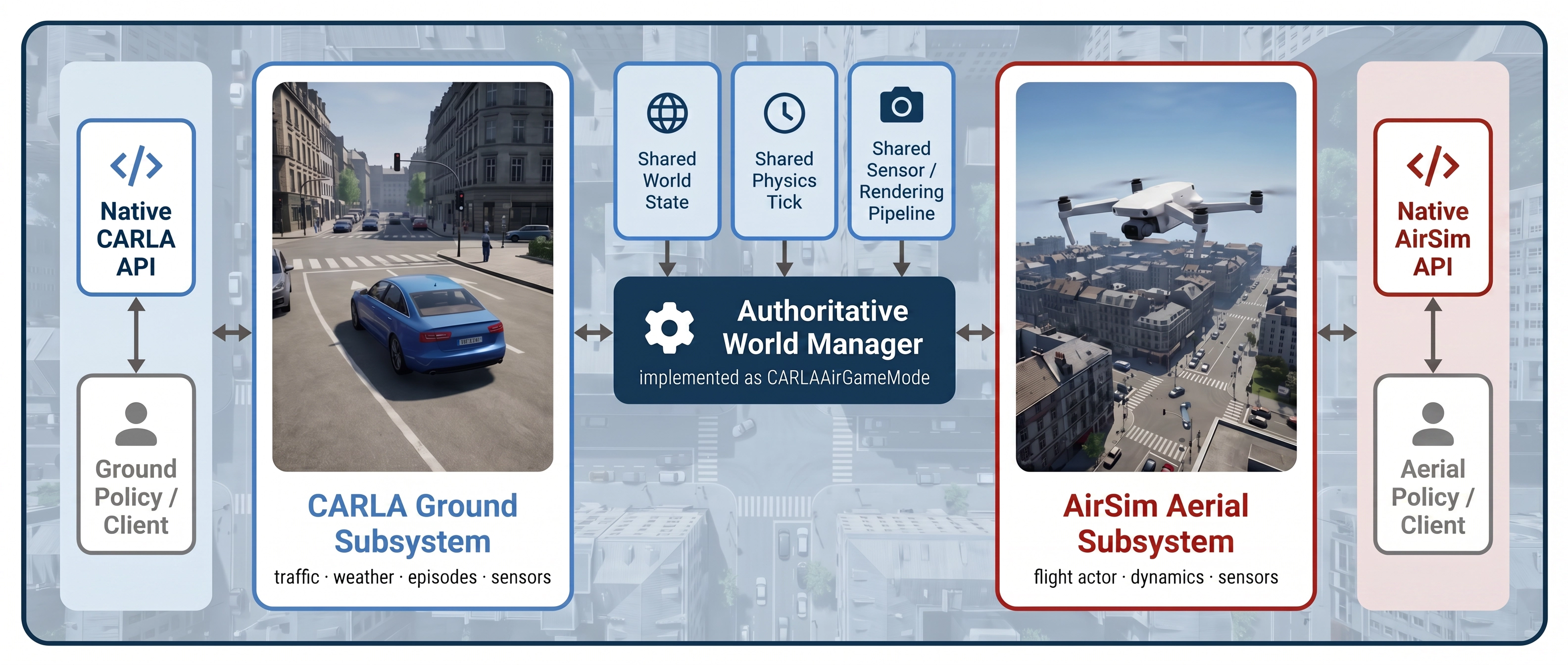}
\caption{%
\textbf{\carlaair{} runtime architecture.}
CARLA and AirSim are embedded in one engine-level runtime with a shared
world state, physics tick, and sensor/rendering pipeline. Native CARLA and AirSim APIs are preserved, while both command streams are resolved inside the same runtime.
}
    \label{fig:runtime_arch}
      \vspace{-0.25cm}
\end{figure*}

\subsection{Runtime Consistency Check}
\label{sec:platform:e0}

Before using \carlaair{} for cooperative VLA evaluation, we verify that 
platform-level synchronization noise is small relative to the cooperation 
effects we aim to measure. As a control, we compare \carlaair{} against 
a standard ROS-bridge configuration in which CARLA and AirSim run as 
separate processes—the architecture used by prior air-ground 
integrations~\cite{chen2026vision}. Both runtimes execute the same 100 matched episodes; the 
only varied factor is the runtime architecture. Full protocol details 
are in Appendix~\ref{app:platform}.

As Table~\ref{tab:e0} shows, the bridge configuration introduces 
non-zero UAV--UGV sensor offsets (mean $12.4$\,ms, max $34$\,ms—roughly 
one third of a $10$\,Hz control cycle), whereas \carlaair{} samples 
matched frames from the same simulation tick and yields $\Delta t = 0$\,ms 
by construction. The cooperation-metric standard deviation drops 
correspondingly from $\sigma = 0.143$ to $\sigma = 0.028$, a $5.1\times$ 
reduction that places \carlaair{}'s noise floor well below the 
cooperation effects reported in Section~\ref{sec:eval}. Extended results 
across heavier sensor loads, higher control rates, and multi-agent 
scenarios in Appendix~\ref{app:platform} show consistent 
$4.2$--$6.3\times$ noise reduction across all settings.

\begin{table}[t]
\centering
\caption{%
  \textbf{Runtime consistency validation.}
  Across 100 matched episodes, the bridge runtime introduces non-zero 
  UAV--UGV sensor offsets and substantially higher variance in the 
  cooperation metric. \carlaair{} eliminates simulation-timestamp 
  mismatch under synchronized tick sampling and reduces measurement 
  noise by $5.1\times$. 
}
\label{tab:e0}
\setlength{\tabcolsep}{7pt}
\renewcommand{\arraystretch}{1.1}
\begin{tabular}{@{}lcc@{}}
\toprule
\textbf{Metric} & \textbf{Bridge runtime} & \textbf{\carlaair{} runtime} \\
\midrule
Mean sensor offset (ms)       & $12.4$  & \textbf{0.0} \\
Std. sensor offset (ms)       & $8.9$   & \textbf{0.0} \\
Max sensor offset (ms)        & $34$    & \textbf{0.0} \\
Std. cooperation metric       & $0.143$ & \textbf{0.028} \\
Measurement-noise reduction   & $1.0\times$ & \textbf{$5.1\times$} \\
\bottomrule
\end{tabular}
  \vspace{-0.25cm}
\end{table}

\begin{figure*}[t]
    \centering
    \includegraphics[width=\textwidth]{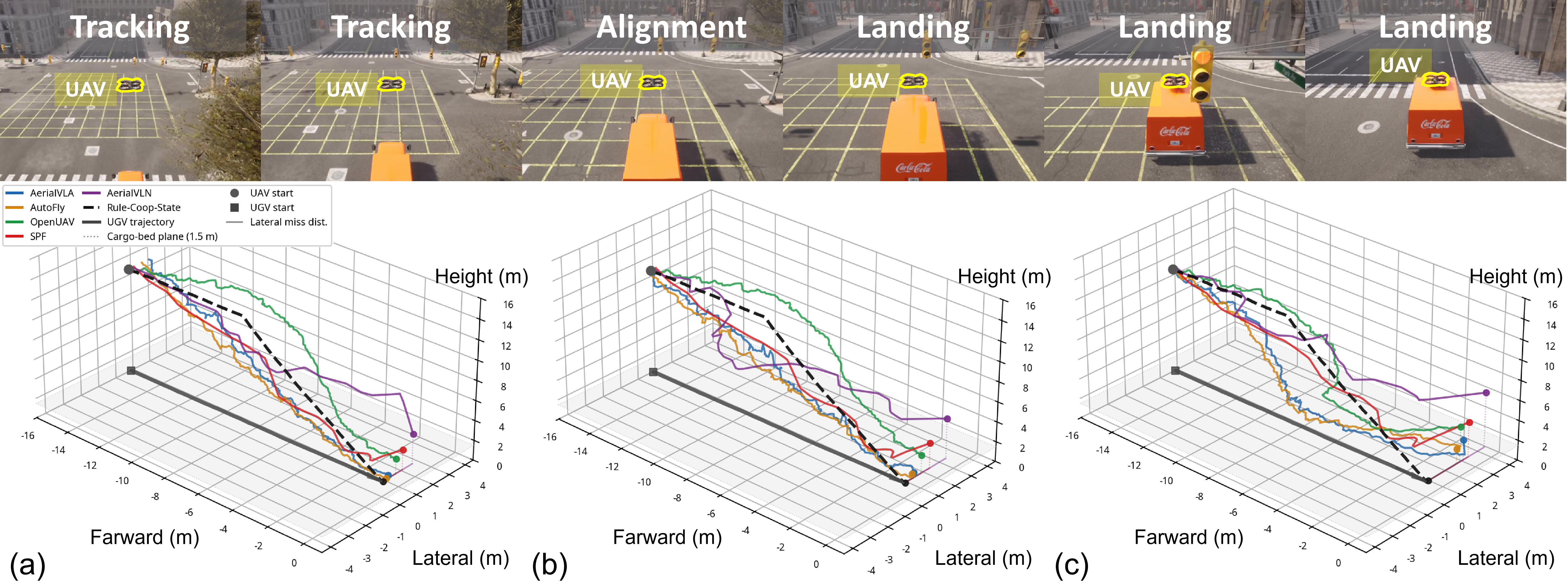}
    \caption{
    \textbf{Landing process trajectory diagnosis.}
    Top: representative visual sequence during moving-platform landing.
    Bottom: 3D UGV and UAV trajectories during moving-platform landing under (a) C0, (b) C1, and (c) C2. Colored curves denote aerial baselines; gray line denotes the UGV trajectory; black dashed line denotes Rule-Coop-State.
    }
    \label{fig:landing_process}
      \vspace{-0.55cm}
\end{figure*}

\section{Diagnostic Evaluation}
\label{sec:eval}

Building on the physically consistent and closed-loop evaluation runtime,
we use \carlaair{} to answer the central question of this paper: can
existing aerial VLA models transfer their single-UAV competence to
cooperative air-ground behavior?

We design two complementary diagnostic tasks. \emph{Cooperative
Moving-Platform Landing} diagnoses action-side coordination: the UAV must
track a moving UGV, align with its rear cargo bed, and land safely.
\emph{Cooperative Occlusion-Recovery Escort} diagnoses perception-side
recovery: the UAV must recover visual contact with a temporarily occluded
UGV using partner-state cues. Both tasks keep the native aerial
VLA interface unchanged: the policy receives visual observations and
language instructions, and outputs only UAV actions.

\subsection{Tasks and Cooperation Protocol}
\label{sec:eval:tasks_protocol}

In Moving-Platform Landing, a UGV truck drives along an urban road while
providing a flat rear cargo bed as the landing surface. The task requires
the UAV to convert visible tracking into aligned descent and touchdown on
a moving platform. In Occlusion-Recovery Escort, the UGV becomes
temporarily occluded by bridges, buildings, or large artifacts; the UAV
must use partner-state cues to re-establish visual contact after the
occlusion.

We evaluate three cooperation modes. C0 uses independent execution with no communication. C1 adds a compact UGV-to-UAV partner-state cue while the UGV follows its predefined behavior. C2, used only for the landing task, additionally passes the magnitude of the UAV's forward-velocity action signal to the UGV's longitudinal controller, providing a naive form of bidirectional UAV-to-UGV action coupling. The signal is derived from each baseline's native action interface (Appendix~\ref{app:eval_details}). Escort uses only C0 and C1. Full prompt templates and controller details are provided in Appendix~\ref{app:eval_details}.

\subsection{Baselines and Metrics}
\label{sec:eval:baselines_metrics}

\begin{table}[t]
\centering
\caption{%
\textbf{Baseline and reference summary.}
}
\label{tab:baseline_summary}
\small
\setlength{\tabcolsep}{4pt}
\renewcommand{\arraystretch}{1.1}
\begin{tabular}{@{}lll@{}}
\toprule
\textbf{Method} & \textbf{Category} & \textbf{Native output} \\
\midrule
\multicolumn{3}{@{}l}{\emph{Zero-shot aerial policies}} \\
AerialVLA~\cite{aerialvla2025}
& End-to-end VLA
& Continuous UAV action \\

OpenFly~\cite{gao2025openfly}
& End-to-end VLA
& Discrete UAV action \\

OpenUAV~\cite{wang2024towards}
& End-to-end VLA
& Continuous UAV action \\

SPF~\cite{spf2025}
& VLM + explicit planning
& UAV flight waypoint \\

AerialVLN~\cite{liu2023aerialvln}
& End-to-end VLN
& Discrete UAV action \\

\midrule
\multicolumn{3}{@{}l}{\emph{State-based cooperative reference}} \\
Rule-Coop-State
& Rule-based cooperation
& UAV/UGV actions \\
\bottomrule
\end{tabular}
  \vspace{-0.25cm}
\end{table}

Table~\ref{tab:baseline_summary} summarizes the evaluated methods,
spanning end-to-end VLA, VLM-based
planning, and traditional aerial VLN. Each baseline is run through its
native low-level wrapper (Appendix~\ref{app:eval_details}); reported
numbers reflect policy-family characteristics rather than
implementation-matched rankings.

Rule-Coop-State is included as a solvability reference, not as a fair
baseline. It uses explicit metric state---UAV--cargo-bed relative pose,
relative velocity, UGV speed, and landing phase---unavailable to VLA
baselines, with deterministic low-latency rules. The gap between
Rule-Coop-State and VLA methods therefore reflects both architectural
and informational differences. We interpret all subsequent results
under this framing.

For landing, we report Tracking Success Rate (TSR) as the single-UAV
primitive score and Landing Success Rate (LSR) as the final cooperative
task score. We define Cooperative Conversion Rate as
$\mathrm{CCR}=\mathrm{LSR}/\max(\mathrm{TSR},\varepsilon)$ with
$\varepsilon=0.05$, and Cooperation Gain as
$\mathrm{CG}(C_k)=\mathrm{LSR}(C_k)-\mathrm{LSR}(C_0)$. For occlusion
recovery, we report Recovery Success Rate (RSR) and Re-acquisition Time
(RAT). We also report Decision Frequency (DF) and Effective Coordination
Latency (ECL). Detailed metric definitions are provided in
Appendix~\ref{app:eval_details}.

\subsection{Results}
\label{sec:eval:results}

\textbf{Moving-platform landing.}
Table~\ref{tab:landing_main} shows that VLA baselines achieve non-trivial
TSR under C0, but their LSR and CCR remain much lower: tracking rarely
converts into successful cooperative landing.

Adding interaction does not close this gap. Under C1, 4 of 5 baselines
exhibit negative CG, but the across-baseline trend is not significant;
under C2, all 5 baselines are negative, a significant trend
($p{=}0.188$ vs.\ $0.031$, sign test). At the single-baseline level,
AerialVLA's CG is statistically negative under both modes, with 95\% CIs
excluding zero: CG(C1) $\in [-0.07, -0.01]$ and CG(C2) $\in [-0.10, -0.04]$.
Naive UAV-to-UGV action coupling under C2 thus introduces unstable
cooperative dynamics rather than stabilizing the task.

Rule-Coop-State reaches LSR $0.42$, indicating that the task is tractable
when explicit state grounding is available. Figure~\ref{fig:landing_process} provides a process-level diagnosis:
VLA baselines often follow the truck but fail to converge smoothly to the
cargo-bed region, while C2 introduces larger oscillations and
Rule-Coop-State produces a more stable descent.

\begin{table*}[t]
\centering
\caption{%
\textbf{Main results on Cooperative Moving-Platform Landing.}
C0/C1/C2: cooperation modes (Section~\ref{sec:eval:tasks_protocol}).
Ref.: state-based cooperative reference.
Values are mean $\pm$ std over 3 seeds $\times$ 50 episodes.
}
\label{tab:landing_main}
\small
\setlength{\tabcolsep}{4.5pt}
\renewcommand{\arraystretch}{1.1}
\begin{tabular}{@{}llcccc@{}}
\toprule
\textbf{Method} & \textbf{Mode}
& \textbf{TSR}$\uparrow$
& \textbf{LSR}$\uparrow$
& \textbf{CCR}$\uparrow$
& \textbf{CG}$\uparrow$ \\
\midrule
AerialVLA & C0 & $0.78{\pm}0.04$ & $0.13{\pm}0.03$ & $0.17{\pm}0.02$ & $0.00$ \\
AerialVLA & C1 & $0.76{\pm}0.04$ & $0.10{\pm}0.03$ & $0.13{\pm}0.02$ & $-0.03{\pm}0.02$ \\
AerialVLA & C2 & $0.70{\pm}0.04$ & $0.06{\pm}0.02$ & $0.09{\pm}0.02$ & $-0.07{\pm}0.02$ \\
\midrule
OpenFly & C0 & $0.81{\pm}0.03$ & $0.14{\pm}0.03$ & $0.17{\pm}0.02$ & $0.00$ \\
OpenFly & C1 & $0.79{\pm}0.04$ & $0.10{\pm}0.03$ & $0.13{\pm}0.02$ & $-0.04{\pm}0.02$ \\
OpenFly & C2 & $0.73{\pm}0.04$ & $0.05{\pm}0.02$ & $0.07{\pm}0.02$ & $-0.09{\pm}0.02$ \\
\midrule
OpenUAV & C0 & $0.74{\pm}0.04$ & $0.09{\pm}0.03$ & $0.12{\pm}0.02$ & $0.00$ \\
OpenUAV & C1 & $0.70{\pm}0.04$ & $0.06{\pm}0.02$ & $0.09{\pm}0.02$ & $-0.03{\pm}0.02$ \\
OpenUAV & C2 & $0.64{\pm}0.04$ & $0.03{\pm}0.02$ & $0.05{\pm}0.02$ & $-0.06{\pm}0.02$ \\
\midrule
SPF & C0 & $0.62{\pm}0.04$ & $0.06{\pm}0.02$ & $0.10{\pm}0.02$ & $0.00$ \\
SPF & C1 & $0.60{\pm}0.04$ & $0.04{\pm}0.02$ & $0.07{\pm}0.02$ & $-0.02{\pm}0.02$ \\
SPF & C2 & $0.55{\pm}0.04$ & $0.02{\pm}0.01$ & $0.04{\pm}0.02$ & $-0.04{\pm}0.02$ \\
\midrule
AerialVLN & C0 & $0.55{\pm}0.03$ & $0.03{\pm}0.02$ & $0.05{\pm}0.02$ & $0.00$ \\
AerialVLN & C1 & $0.51{\pm}0.03$ & $0.02{\pm}0.01$ & $0.04{\pm}0.01$ & $-0.01{\pm}0.01$ \\
AerialVLN & C2 & $0.47{\pm}0.03$ & $0.01{\pm}0.01$ & $0.02{\pm}0.01$ & $-0.02{\pm}0.01$ \\
\midrule
\multicolumn{6}{@{}l}{\emph{State-based cooperative reference}} \\
Rule-Coop-State & Ref. & $\mathbf{0.84}{\pm}0.03$ & $\mathbf{0.42}{\pm}0.03$ & $\mathbf{0.50}{\pm}0.03$ & --- \\
\bottomrule
\end{tabular}
  \vspace{-0.8cm}
\end{table*}

\begin{figure}[t]
    \centering
    \includegraphics[width=\linewidth]{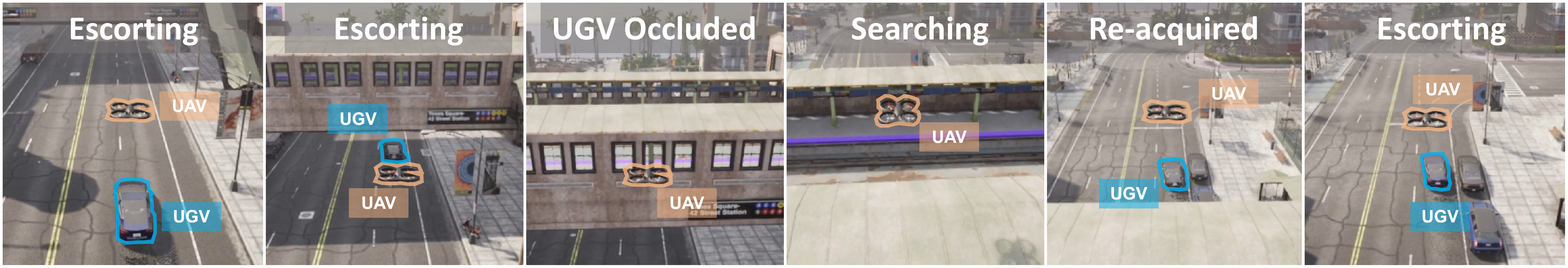}
\caption{
\textbf{Cooperative Occlusion-Recovery Escort.}
The UAV escorts the UGV, loses visual contact under temporary occlusion,
searches using partner-state cues, and re-acquires the UGV before resuming
escort.
}
    \label{fig:occlusion_process}
      \vspace{-0.65cm}
\end{figure}

\begin{table}[t]
\centering
\caption{%
\textbf{Results on Cooperative Occlusion-Recovery Escort.}
RSR: visual contact recovered ($\mathrm{IoU} \geq 0.15$) within 15\,s
of occlusion onset. RAT: re-acquisition time (s); capped at 15\,s.
Ref.: state-based cooperative reference.
Values are mean $\pm$ std over 3 seeds $\times$ 50 episodes.
}
\label{tab:occlusion_results}
\small
\setlength{\tabcolsep}{5pt}
\renewcommand{\arraystretch}{1.1}
\begin{tabular}{@{}llcc@{}}
\toprule
\textbf{Method} & \textbf{Mode}
& \textbf{RSR}$\uparrow$
& \textbf{RAT (s)}$\downarrow$ \\
\midrule
AerialVLA & C0 & $0.56{\pm}0.04$ & $4.8{\pm}0.3$ \\
AerialVLA & C1 & $0.58{\pm}0.04$ & $4.6{\pm}0.3$ \\
\midrule
OpenFly & C0 & $0.59{\pm}0.04$ & $4.5{\pm}0.3$ \\
OpenFly & C1 & $0.57{\pm}0.04$ & $4.7{\pm}0.3$ \\
\midrule
OpenUAV & C0 & $0.48{\pm}0.04$ & $6.2{\pm}0.4$ \\
OpenUAV & C1 & $0.44{\pm}0.04$ & $6.7{\pm}0.4$ \\
\midrule
SPF & C0 & $0.46{\pm}0.04$ & $6.5{\pm}0.4$ \\
SPF & C1 & $0.54{\pm}0.04$ & $5.4{\pm}0.3$ \\
\midrule
AerialVLN & C0 & $0.41{\pm}0.03$ & $7.2{\pm}0.4$ \\
AerialVLN & C1 & $0.40{\pm}0.03$ & $7.4{\pm}0.4$ \\
\midrule
\multicolumn{4}{@{}l}{\emph{State-based cooperative reference}} \\
Rule-Coop-State & Ref. & $\mathbf{0.86}{\pm}0.03$ & $\mathbf{1.8}{\pm}0.2$ \\
\bottomrule
\end{tabular}
  \vspace{-0.8cm}
\end{table}

\begin{table}[t]
\centering
\caption{%
\textbf{Prompt-format ablation (landing and escort).}
Cue variants: C1-Sem (default semantic), C1-Num (structured numeric),
C1-Noisy (corrupted), C1-Oracle-Bearing (ground-truth range / azimuth /
elevation). Values are mean $\pm$ std over 3 seeds $\times$ 50 episodes.
Full prompt templates in Appendix~\ref{app:eval_details}.
}
\label{tab:prompt_ablation_results}
\small
\setlength{\tabcolsep}{3.5pt}
\renewcommand{\arraystretch}{1.1}
\begin{tabular}{@{}llccccc@{}}
\toprule
\textbf{Method} & \textbf{Setting}
& \textbf{LSR}$\uparrow$
& \textbf{CCR}$\uparrow$
& \textbf{CG}$\uparrow$
& \textbf{RSR}$\uparrow$
& \textbf{RAT}$\downarrow$ \\
\midrule
AerialVLA & C0               & $0.13{\pm}0.03$ & $0.17{\pm}0.02$ & $0.00$           & $0.56{\pm}0.04$ & $4.8{\pm}0.3$ \\
AerialVLA & C1-Sem           & $0.10{\pm}0.03$ & $0.13{\pm}0.02$ & $-0.03{\pm}0.02$ & $0.58{\pm}0.04$ & $4.6{\pm}0.3$ \\
AerialVLA & C1-Num           & $0.08{\pm}0.02$ & $0.11{\pm}0.02$ & $-0.05{\pm}0.02$ & $0.52{\pm}0.04$ & $5.2{\pm}0.3$ \\
AerialVLA & C1-Noisy         & $0.05{\pm}0.02$ & $0.07{\pm}0.02$ & $-0.08{\pm}0.02$ & $0.43{\pm}0.04$ & $6.4{\pm}0.4$ \\
AerialVLA & C1-Oracle-Bearing& $0.14{\pm}0.03$ & $0.18{\pm}0.02$ & $+0.01{\pm}0.02$ & $0.62{\pm}0.04$ & $4.3{\pm}0.3$ \\
\midrule
OpenFly & C0               & $0.14{\pm}0.03$ & $0.17{\pm}0.02$ & $0.00$           & $0.59{\pm}0.04$ & $4.5{\pm}0.3$ \\
OpenFly & C1-Sem           & $0.10{\pm}0.03$ & $0.13{\pm}0.02$ & $-0.04{\pm}0.02$ & $0.57{\pm}0.04$ & $4.7{\pm}0.3$ \\
OpenFly & C1-Num           & $0.07{\pm}0.02$ & $0.09{\pm}0.02$ & $-0.07{\pm}0.02$ & $0.50{\pm}0.04$ & $5.5{\pm}0.3$ \\
OpenFly & C1-Noisy         & $0.04{\pm}0.02$ & $0.06{\pm}0.02$ & $-0.10{\pm}0.02$ & $0.41{\pm}0.04$ & $6.8{\pm}0.4$ \\
OpenFly & C1-Oracle-Bearing& $0.13{\pm}0.03$ & $0.16{\pm}0.02$ & $-0.01{\pm}0.02$ & $0.61{\pm}0.04$ & $4.4{\pm}0.3$ \\
\bottomrule
  \vspace{-0.95cm}
\end{tabular}

\end{table}

\textbf{Timing.}
Heavier planning-based methods have lower decision frequency and larger
effective coordination latency, making C2 feedback harder to stabilize
(Appendix~\ref{app:eval_details}, Table~\ref{tab:timing_app}). Yet latency alone does not explain the failure: faster methods such as AerialVLN still show poor cooperative conversion. The degradation is therefore both temporal and structural.

\textbf{Prompt-format ablation.}
Table~\ref{tab:prompt_ablation_results} shows a consistent pattern across
both tasks: semantic, numeric, and noisy cue formats all fail to close
the cooperation gap. Even oracle geometric cues (C1-Oracle-Bearing)
provide only marginal LSR gains, with CGs within seed-level variability.
Under this text-prompt interface, the bottleneck is action grounding
rather than cue format or completeness.

\textbf{Occlusion recovery.}
Figure~\ref{fig:occlusion_process} illustrates the occlusion-recovery
escort task. Table~\ref{tab:occlusion_results} shows that VLA baselines do not
consistently benefit from C1 partner-state cues; for 4 of 5 baselines,
the C0 vs.\ C1 RSR difference is within seed-level variability. SPF
shows a meaningful RSR gain (+$0.08$), likely because the cue
(\emph{forward-right side}) aligns with its native waypoint output
space; the same cue provides no benefit in landing, where it must be
decomposed into continuous descent control. Rule-Coop-State reaches
RSR $0.86$.

\textbf{Summary diagnosis.}
Across both tasks, current zero-shot aerial VLA policies fail to
translate partner-state cues into cooperative action: tracking does
not become coordinated landing, and partner-state cues do not become
recovery behavior. The failure is not in cue design or cue completeness:
across-baseline statistical evidence, prompt-format ablations, and
oracle bearing cues all converge on the same
attribution---the missing mechanism is partner-aware action grounding
within the autoregressive text-prompt interface itself.
Rule-Coop-State, with explicit metric state and rule-based coordination,
substantially outperforms VLA baselines on both tasks, marking the gap as
architectural rather than fundamental to the cooperation problem.

\section{Discussion and Conclusion}
\label{sec:discussion}
\textbf{Interaction is not cooperation.}
Adding cross-agent information does not automatically produce cooperation.
C1 and C2 provide more partner state and action feedback, yet performance
can degrade. The missing component is not information, but a mechanism
that maps partner state into the model's own action for a shared task.

\textbf{Timing amplifies the failure.}
Larger effective coordination latency makes bidirectional feedback harder
to stabilize, but latency alone does not explain the failure: faster
aerial VLAs still show poor cooperative conversion, suggesting that the
degradation is structural rather than purely temporal.

\textbf{Implications.}
Cooperative VLA likely requires explicit partner-state modeling, shared
task-phase representation, team-level objectives, and low-latency
coordination modules.

\textbf{Limitations.}
Our study has five principal boundaries:
\textit{(i)} simulation only;
\textit{(ii)} the cooperative burden falls primarily on the UAV side,
with UGV-side cooperative planning unstudied;
\textit{(iii)} information asymmetry between Rule-Coop-State and VLA
baselines;
\textit{(iv)} zero-shot VLA evaluation without cooperation-aware
fine-tuning;
\textit{(v)} a limited set of occlusion types, with learned cooperation
policies and non-VLA baselines left for future work.
The sim-to-real effect remains empirically open.

\textbf{Broader impact.}
\carlaair{} could be repurposed for aerial tracking or surveillance, and
safety-critical or privacy-sensitive deployments require independent
ethical review. Our results caution against deploying zero-shot aerial
VLA systems in cooperative scenarios with moving ground partners without
cooperation-aware design and closed-loop validation.

\textbf{Conclusion.}
We presented \carlaair{}, a single-process air-ground evaluation
environment for closed-loop UAV--UGV cooperation, and used it to diagnose
whether single-UAV aerial VLA competence transfers to cooperative
air-ground behavior. Across moving-platform landing and occlusion-recovery
escort, current zero-shot aerial VLA models can track or follow a ground
partner but fail to convert this ability into stable cooperation:
partner-state prompting helps inconsistently, and naive bidirectional
interaction can amplify errors. Prompt-format ablations and
oracle-bearing cues show that the bottleneck is not cue design but
partner-state grounding into action within a text-prompt interface.

\bibliographystyle{unsrtnat}
\bibliography{new_references}

\newpage
\clearpage
\appendix

\section*{Appendix}
\label{app:overview}

This appendix provides additional platform and evaluation details that
support the main paper. Appendix~\ref{app:platform} describes the
\carlaair{} runtime design, coordinate-frame unification, sensing support,
software stack, and runtime consistency validation.
Appendix~\ref{app:eval_details} provides task settings, cooperation modes,
prompt templates, metric definitions, baseline adaptations, and timing
statistics.

\section{\carlaair{} Platform Details}
\label{app:platform}

\begin{figure}[H]
    \centering
    \includegraphics[width=0.95\linewidth]{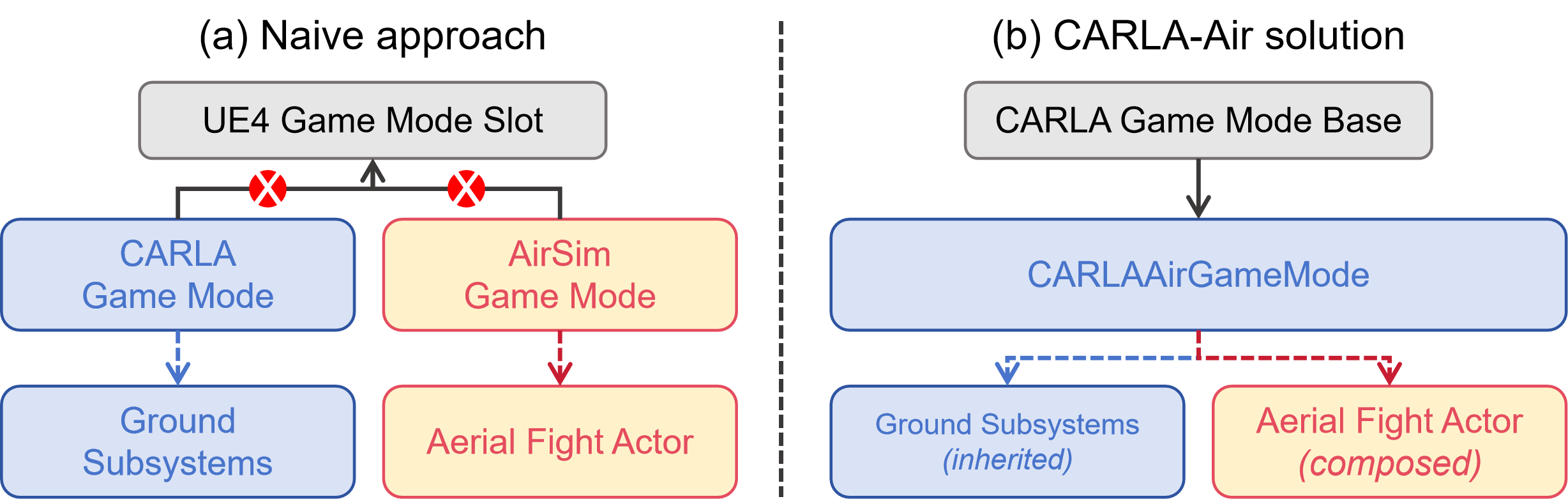}
    \caption{
    \textbf{Resolving the single-GameMode constraint.}
    A naive merge of CARLA and AirSim creates competing Unreal Engine
    GameMode controllers. \carlaair{} preserves CARLA as the authoritative
    world manager through \texttt{CARLAAirGameMode} and composes the AirSim
    flight actor as a regular world entity, enabling single-process
    air-ground simulation without competing runtime controllers.
    }
    \label{fig:app_gamemode_resolution}
\end{figure}

This appendix provides implementation details for \carlaair{}, including
coordinate-frame unification, sensor and API support, software versions,
source modifications, and runtime requirements.
Figure~\ref{fig:app_gamemode_resolution} illustrates the core integration
mechanism: \carlaair{} resolves the single-GameMode constraint by keeping
CARLA as the authoritative world manager and composing the AirSim aerial
subsystem as an actor-level component.

\subsection{Coordinate Frame Unification and Single-Tick Execution}
\label{app:coord_frame}

\begin{figure}[t]
    \centering
    \includegraphics[width=0.9\linewidth]{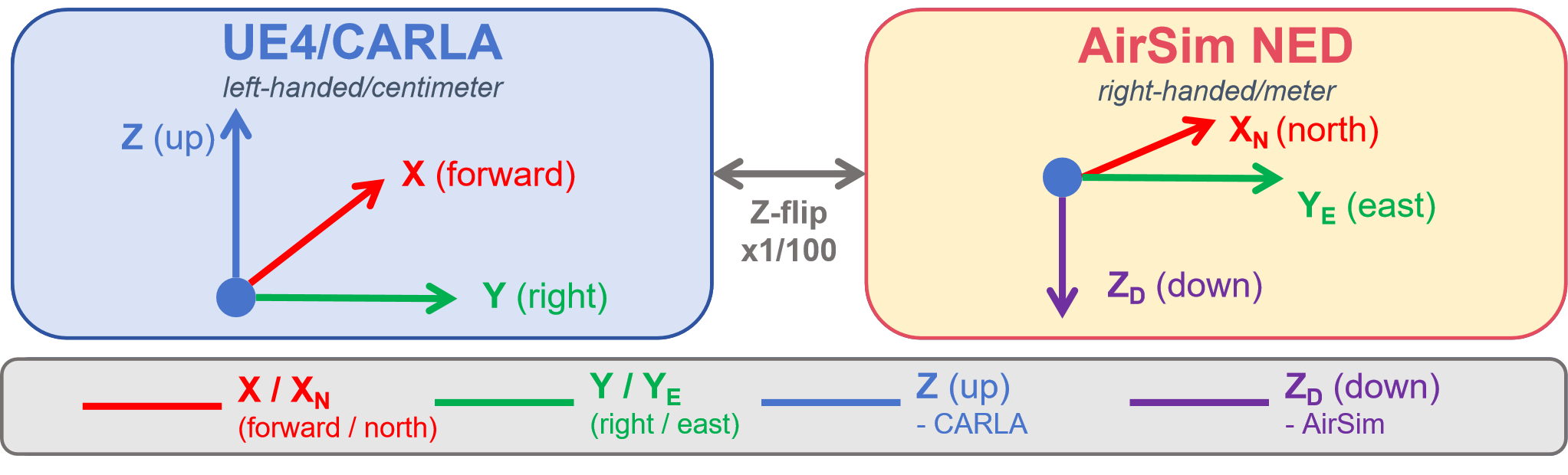}
    \caption{
    \textbf{Coordinate-frame alignment.}
    CARLA uses the Unreal Engine frame in centimetres with $Z$-up, while
    AirSim uses a metre-scale NED frame with $Z$-down. \carlaair{} applies
    a deterministic scale conversion and $Z$-axis sign flip to express UAV
    and UGV states in one metric frame.
    }
    \label{fig:app_coordinate_frames}
\end{figure}

All cross-agent states, observations, and cooperation metrics are expressed
in a unified metric frame. CARLA uses a left-handed Unreal Engine frame
(centimetres, $Z$-up), while AirSim adopts a NED frame (metres,
$Z$-down). As illustrated in Figure~\ref{fig:app_coordinate_frames}, we apply a deterministic transformation at every simulation tick:
\begin{equation}
  \mathbf{p}_{\mathrm{NED}}
  = \tfrac{1}{100}
  \begin{pmatrix} p_x - o_x \\ p_y - o_y \\ -(p_z - o_z) \end{pmatrix},
  \quad
  q_{\mathrm{NED}} =
  \frac{(w,\, q_x,\, q_y,\, -q_z)}
       {\|(w,\, q_x,\, q_y,\, -q_z)\|},
  \label{eq:coord_transform}
\end{equation}
where $(o_x,o_y,o_z)$ is the AirSim origin in Unreal Engine coordinates.
This transformation ensures that UAV states, UGV states, relative poses,
and cooperation metrics are evaluated in the same metric frame.

Algorithm~\ref{alg:single_tick} summarizes the per-tick execution flow:
ground and aerial commands are applied to the shared world, physics is
advanced once, and UAV/UGV sensors are sampled from the same updated
state. Rendering is forced to complete within the same tick via
\texttt{FlushRenderingCommands()}, so all sensor outputs are computed
from a single physics state before the tick returns.

\begin{algorithm}[H]
\caption{Single-tick execution in \carlaair{}}
\label{alg:single_tick}
\begin{algorithmic}[1]
\REQUIRE World state $\mathcal{W}_t$, UGV action $a^G_t$, UAV action $a^A_t$
\ENSURE Updated world $\mathcal{W}_{t+1}$, synchronized observations $o^G_{t+1}, o^A_{t+1}$
\STATE Apply UGV command $a^G_t$ through the CARLA control interface
\STATE Apply UAV command $a^A_t$ through the AirSim control interface
\STATE $\mathcal{W}_{t+1} \leftarrow \textsc{PhysicsStep}(\mathcal{W}_t)$
\STATE Render and sample sensors from $\mathcal{W}_{t+1}$
\STATE $o^G_{t+1} \leftarrow \textsc{SampleSensors}(\mathrm{UGV}, \mathcal{W}_{t+1})$
\STATE $o^A_{t+1} \leftarrow \textsc{SampleSensors}(\mathrm{UAV}, \mathcal{W}_{t+1})$
\STATE \textbf{return} $\mathcal{W}_{t+1}, o^G_{t+1}, o^A_{t+1}$
\end{algorithmic}
\end{algorithm}
\subsection{Sensors and Native APIs}
\label{app:sensors}

\begin{figure*}[t]
    \centering
    \includegraphics[width=\textwidth]{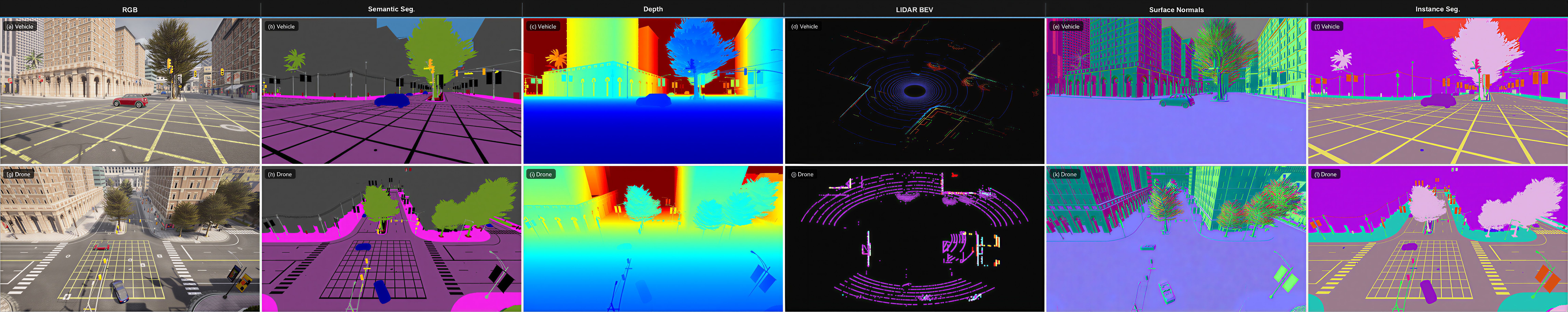}
    \caption{
    \textbf{Example of synchronized aerial-ground sensing.}
    Vehicle-side and UAV-side sensor streams are sampled from the same
    simulation tick in \carlaair{}. The top row shows vehicle-side
    modalities and the bottom row shows UAV-side modalities, including RGB,
    semantic segmentation, depth, LiDAR BEV, surface normals, and instance
    segmentation. This provides paired multi-modal observations without
    cross-process timestamp alignment.
    }
    \label{fig:app_sync_sensors}
\end{figure*}

Table~\ref{tab:sensors} lists the sensor modalities supported by
\carlaair{}. All enabled sensors are sampled at the shared simulation
tick, so cross-agent correspondence does not require interpolation or
extrapolation. CARLA and AirSim retain their original Python APIs and ROS\,2
interfaces; both command streams are resolved inside the same Unreal
Engine runtime, so existing client code continues to operate without
modification. Figure~\ref{fig:app_sync_sensors} visualizes this same-tick
aerial-ground sensing setup.

\begin{table}[t]
\centering
\caption{Sensor modalities available in \carlaair{}. UGV and UAV columns
indicate availability on each platform.}
\label{tab:sensors}
\small
\setlength{\tabcolsep}{5pt}
\renewcommand{\arraystretch}{1.1}
\begin{tabular}{@{}llcc@{}}
\toprule
\textbf{\#} & \textbf{Sensor} & \textbf{UGV} & \textbf{UAV} \\
\midrule
1  & RGB camera (forward)             & \checkmark & \checkmark \\
2  & RGB camera (downward)            & ---        & \checkmark \\
3  & RGB camera (wide-angle)          & \checkmark & \checkmark \\
4  & Depth camera                     & \checkmark & \checkmark \\
5  & Semantic segmentation camera     & \checkmark & \checkmark \\
6  & Instance segmentation camera     & \checkmark & \checkmark \\
7  & Surface normal camera            & \checkmark & \checkmark \\
8  & LiDAR                            & \checkmark & \checkmark \\
9  & Radar                            & \checkmark & ---        \\
10 & IMU                              & \checkmark & \checkmark \\
11 & GNSS                             & \checkmark & \checkmark \\
12 & Barometer                        & ---        & \checkmark \\
13 & Magnetometer                     & ---        & \checkmark \\
14 & Optical flow camera              & \checkmark & \checkmark \\
15 & Event camera                     & \checkmark & \checkmark \\
16 & Collision sensor                 & \checkmark & \checkmark \\
17 & Lane invasion sensor             & \checkmark & ---        \\
18 & Obstacle distance sensor         & \checkmark & \checkmark \\
\bottomrule
\end{tabular}
\end{table}

\subsection{Software and Hardware}
\label{app:software_hardware}

Table~\ref{tab:software_versions} summarizes the software stack used in
our experiments. All builds and experiments are run on a single workstation with an Intel Core Ultra 9 275HX (24C/24T) CPU, an NVIDIA RTX 5090 Laptop GPU (24 GB),
and 128 GB RAM under Ubuntu 22.04. Runtime evaluation
requires a GPU with sufficient memory for the selected VLA baseline.

\begin{table}[t]
\centering
\caption{Software stack used in our experiments.}
\label{tab:software_versions}
\small
\setlength{\tabcolsep}{8pt}
\renewcommand{\arraystretch}{1.1}
\begin{tabular}{@{}ll@{}}
\toprule
\textbf{Component} & \textbf{Version} \\
\midrule
Unreal Engine          & 4.26.2 \\
CARLA                  & 0.9.16 \\
AirSim                 & 1.7.0 \\
Python                 & 3.8.18 \\
PyTorch                & 2.1.2 (CUDA 11.8) \\
ROS 2                  & Humble Hawksbill \\
Ubuntu                 & 22.04 LTS \\
\bottomrule
\end{tabular}
\end{table}

\noindent\textbf{Note on AirSim.}
The original open-source AirSim project is no longer actively maintained
by Microsoft. We build on Colosseum~\cite{colosseum}, a community-maintained
fork that preserves the AirSim API surface and remains compatible with
Unreal Engine.

\subsection{Runtime Consistency Validation}
\label{app:platform:e0_protocol}

This section provides the experimental details underlying the runtime
consistency check reported in Section~\ref{sec:platform:e0}
(Table~\ref{tab:e0}), and extends the validation to additional sensor
configurations, control frequencies, and agent counts.

\paragraph{Setup.}
The UGV drives along a predefined route in Town10 at constant speed,
while the UAV is driven by a position P-controller to follow the UGV at
a fixed relative offset; the underlying flight control uses AirSim's
built-in cascaded PID module (the AirSim default
configuration~\citep{airsim2017}).

\paragraph{Comparison configurations.}
Under the bridge runtime, CARLA and AirSim run as separate processes
and exchange messages via ROS\,2 at 10\,Hz---the architecture commonly
used by prior air-ground integrations~\citep{chen2026vision}.
Under the \carlaair{} runtime, both simulators share a single
simulation tick. Both runtimes execute the same 100 episodes with
matched random seeds, identical spawn points, and the same follow
controller; the only varied factor is the runtime architecture.

\paragraph{Measurements.}
\emph{Sensor offset} is the per-tick timestamp difference
$|\tau_{\mathrm{UAV}} - \tau_{\mathrm{UGV}}|$ (ms) between matched UAV
and UGV sensor frames. \emph{Follow-error std} is the per-episode mean
of the Euclidean distance (in metres) between the UAV's actual relative
pose and its target relative pose during the steady-state segment of
the episode, taken as the standard deviation across the 100 episodes;
this is the $\sigma$ reported in Table~\ref{tab:e0}.
\emph{Wall-clock jitter} (used in Table~\ref{tab:stress_test}) is the
standard deviation of OS frame-delivery timestamps after the physics
tick returns, and arises from GPU flush overhead.
The reported $5.1\times$ noise reduction has a $95\%$ confidence
interval of $[3.4\times, 7.7\times]$ (F-distribution,
$n_1 = n_2 = 100$).

\paragraph{Stress-test extension.}
Table~\ref{tab:stress_test} extends this protocol to heavier sensor
payloads, higher control rates, multi-agent settings, and dense traffic.
\carlaair{} maintains $\Delta t = 0$\,ms by construction in all
configurations, with $\sigma$ ratios in the
$4.2\times$--$6.3\times$ range; remaining wall-clock jitter
(4--12\,ms P95) is substantially lower than bridge timing noise.

\begin{table}[t]
\centering
\caption{%
\textbf{Runtime stress test.}
Bridge timestamp offset (mean/P95, ms), wall-clock delivery jitter (P95, ms),
and metric noise reduction ($\sigma$ ratio = bridge\,/\,\carlaair{} cooperation-metric std,
mean $\pm$ std over 5 independent 100-episode runs).
\carlaair{} maintains zero simulation-timestamp offset in all settings.
}
\label{tab:stress_test}
\small
\setlength{\tabcolsep}{4pt}
\renewcommand{\arraystretch}{1.1}
\begin{tabular}{@{}lcccc@{}}
\toprule
\makecell{\textbf{Setting}\\\textbf{(agents, sensors, Hz)}}
& \makecell{\textbf{Bridge offset}\\\textbf{mean/P95 (ms)}}
& \makecell{\textbf{\carlaair{}}\\\textbf{offset (ms)}}
& \makecell{\textbf{WC jitter}\\\textbf{P95 (ms)}}
& \makecell{\textbf{$\sigma$}\\\textbf{ratio}} \\
\midrule
Easy-RGB (1U+1G, RGB, 10\,Hz)             & 12.4/34 & \textbf{0.0} & 4  & $5.1{\pm}0.4\times$ \\
Multi-sensor (1U+1G, RGB+D+L, 10\,Hz)    & 15.2/41 & \textbf{0.0} & 7  & $4.8{\pm}0.4\times$ \\
High-rate (1U+1G, RGB, 30\,Hz)            & 18.7/52 & \textbf{0.0} & 9  & $6.3{\pm}0.5\times$ \\
Multi-agent (2U+2G, RGB+L, 10\,Hz)        & 22.1/64 & \textbf{0.0} & 12 & $4.2{\pm}0.4\times$ \\
Dense-traffic (1U+1G, RGB, 10\,Hz)        & 13.1/37 & \textbf{0.0} & 5  & $5.0{\pm}0.4\times$ \\
\bottomrule
\end{tabular}
\end{table}

\subsection{Source Modification Summary}
\label{app:source_mods}

The \carlaair{} integration modifies only a small number of files relative
to the CARLA upstream codebase:
\begin{itemize}[leftmargin=*]
  \item \texttt{CarlaUE4GameMode.h}: adds the AirSim flight actor
  declaration and composition pointer.
  \item \texttt{CarlaUE4GameMode.cpp}: instantiates the AirSim actor and
  synchronizes it with the CARLA world lifecycle.
  \item \texttt{CarlaUE4.Build.cs}: adds the AirSim module dependency.
\end{itemize}
The integration preserves the native CARLA and AirSim client-facing APIs.

\section{Additional Diagnostic Evaluation Details}
\label{app:eval_details}

This appendix provides details omitted from the main diagnostic evaluation
section, including task settings, cooperation modes, prompt templates,
metric definitions, baseline adaptations, and evaluation protocol.

\subsection{Task Details}
\label{app:task_details}

\paragraph{Cooperative Moving-Platform Landing.}
A UGV truck drives along an urban road while providing a flat rear cargo
bed as the landing surface. The UAV receives the instruction:
\emph{Follow the moving truck, align above its rear cargo bed, and land
safely.} The task consists of tracking, alignment, and landing, with a
60\,s episode time limit. It is successful only when the UAV lands on the
rear cargo bed without collision, side impact, or hard landing.

\paragraph{Cooperative Occlusion-Recovery Escort.}
A UGV drives along an urban route and becomes temporarily occluded by
bridges, buildings, or large artifacts. The UAV must escort the UGV and
recover visual contact after the target becomes invisible. Each escort
episode has a 90\,s time limit. The C1 cue describes the UGV's motion
intent and expected reappearance direction, while the VLA policy still
outputs only UAV actions.

\subsection{Cooperation Modes}
\label{app:coop_modes}

\paragraph{C0: Independent execution.}
The UAV and UGV do not communicate. The UGV follows a predefined speed
profile or route, while the UAV relies only on onboard RGB observations and
the task instruction.

\paragraph{C1: UGV-to-UAV semantic prompting.}
The UGV provides a compact semantic cue to the UAV. For landing, the cue
describes the relative direction to the cargo bed, coarse truck motion, and
landing phase. For occlusion recovery, it describes the occlusion status,
motion intent, and expected reappearance direction. The UAV still outputs
only native UAV actions.

\paragraph{C2: Bidirectional UAV-to-UGV action coupling.}
C2 is used only for Moving-Platform Landing. The UAV receives the same
semantic cue as in C1. The magnitude of the UAV's commanded forward
velocity is passed directly to a fixed UGV longitudinal controller, with
no intermediate phase decoder or learned mapping:
\begin{equation}
  v_{\mathrm{UGV}}(t)
  = v_0 \cdot
    \mathrm{clip}\!\left(
      \frac{\| v_{\mathrm{UAV}}^{\mathrm{fwd}}(t) \|}{v_{\mathrm{ref}}},
      \; 0.5,\; 1.5
    \right),
  \label{eq:c2_controller}
\end{equation}
where $v_0 = 4.0$\,m/s is the nominal UGV speed and
$v_{\mathrm{ref}} = 2.0$\,m/s is a reference scaling constant. The
$\mathrm{clip}(\cdot, 0.5, 1.5)$ operator bounds the multiplicative
factor to $[0.5\times, 1.5\times]$ to prevent extreme values. The
controller modulates only longitudinal speed; the UGV heading and route
remain unchanged. The update frequency matches the UAV decision
frequency. For baselines whose native output is not a continuous velocity
vector (OpenFly, AerialVLN), $\| v_{\mathrm{UAV}}^{\mathrm{fwd}}(t) \|$
is obtained from the realized UAV forward velocity in the simulator at
the same tick. For waypoint-output baselines (SPF, OpenUAV), it is
computed as commanded waypoint displacement divided by the inference
period. The updated UGV state is then fed back to the UAV prompt for the
next step. No additional VLA output head or UGV steering command is
introduced; the protocol is intentionally a naive form of bidirectional
action coupling.

\subsection{Prompt Templates}
\label{app:prompts}

Table~\ref{tab:app_prompt_protocol} lists the full prompt templates used
across tasks and cooperation modes. All VLA baselines receive the same
base instruction within each task, with mode-specific assistant hints
appended for C1 and C2.

\begin{table*}[t]
\centering
\caption{%
\textbf{Full prompt protocol.}
All VLA baselines receive the same base instruction within each task. C1
uses semantic partner-state prompts. C2 uses the same UAV-side prompt as
C1 and only enables the UGV-side longitudinal response to UAV action.
}
\label{tab:app_prompt_protocol}
\small
\setlength{\tabcolsep}{4pt}
\renewcommand{\arraystretch}{1.15}
\begin{tabular}{@{}p{0.12\textwidth}p{0.07\textwidth}p{0.22\textwidth}p{0.51\textwidth}@{}}
\toprule
\textbf{Task} & \textbf{Mode} & \textbf{Interaction} & \textbf{UAV prompt example} \\
\midrule
Landing
& C0
& No communication
& \emph{Follow the moving truck, keep it in view, align above the flat rear cargo bed, and land safely on the cargo bed.} \\
\midrule
Landing
& C1
& Semantic state cue
& \emph{Follow the moving truck, keep it in view, align above the flat rear cargo bed, and land safely on the cargo bed. Assistant hint: the cargo bed is forward-left. The truck is moving slowly. Current phase: approach. Use the hint only to choose your next UAV action.} \\
\midrule
Landing
& C2
& Same UAV prompt as C1; UGV speed responds to UAV forward velocity
& \emph{Follow the moving truck, keep it in view, align above the flat rear cargo bed, and land safely on the cargo bed. Assistant hint: the cargo bed is forward-left. The truck is moving slowly. Current phase: approach. Use the hint only to choose your next UAV action.} \\
\midrule
Escort
& C0
& No communication
& \emph{Follow the moving truck and keep it in view.} \\
\midrule
Escort
& C1
& Occlusion-recovery cue
& \emph{Follow the moving truck and keep it in view. Assistant hint: the truck is temporarily occluded by the bridge. The truck continues forward and will reappear on the forward-right side. Current phase: occlusion recovery. Use the hint only to recover visual contact.} \\
\midrule
Landing
& C1-Oracle-Bearing
& Oracle geometric cue
& \emph{Follow the moving truck, keep it in view, align above the flat rear cargo bed, and land safely. State update: cargo bed at bearing 312°, range 6.2\,m, elevation $-$8°. Phase: approach. Use the state update only to choose your next UAV action.} \\
\bottomrule
\end{tabular}
\end{table*}

\subsection{Metric Details}
\label{app:metric_details}

\paragraph{Tracking Success Rate.}
Tracking Success Rate (TSR) measures whether the target truck remains
inside the UAV camera view for at least $K=3$\,s of cumulative time
before task termination. It evaluates the single-UAV tracking primitive
rather than final task success.

\paragraph{Landing Success Rate.}
Landing Success Rate (LSR) measures the fraction of episodes in which the
UAV lands on the rear cargo bed within the 60\,s time limit and remains
stable after touchdown, defined as no further displacement exceeding
0.3\,m within 2\,s of first contact.

\paragraph{Cooperative Conversion Rate.}
Cooperative Conversion Rate (CCR) measures whether single-UAV tracking
becomes cooperative landing:
\[
\mathrm{CCR} =
\frac{\mathrm{LSR}}{\max(\mathrm{TSR}, \varepsilon)} .
\]
We set $\varepsilon = 0.05$; substituting $\varepsilon \in \{0.01, 0.05, 0.10\}$
changes CCR by at most 0.01 across all reported conditions, as all baselines
achieve $\mathrm{TSR} \geq 0.55$ under C0.
A high TSR with low CCR indicates that the UAV can track the moving
platform but cannot convert tracking into landing.

\paragraph{Cooperation Gain.}
Cooperation Gain (CG) measures the change in LSR relative to independent
execution:
\[
\mathrm{CG}(C_k) =
\mathrm{LSR}(C_k) - \mathrm{LSR}(C_0).
\]
Positive CG means that the cooperation mode improves over C0, while
negative CG indicates degradation.

\paragraph{Occlusion-recovery metrics.}
Recovery Success Rate (RSR) measures whether the UAV recovers visual
contact with the UGV within 15\,s of occlusion onset; success requires
$\mathrm{IoU} \geq 0.15$ between the UAV camera view and the UGV
bounding box, sustained for at least 0.5\,s.
Re-acquisition Time (RAT) is measured from occlusion onset to the first
frame satisfying the IoU threshold; it is capped at 15\,s for episodes
with no recovery.
Each escort episode contains 1--3 occlusion events sampled from three
geometry types (bridge underpass: 40\%, building: 35\%,
large artifacts: 25\%), with occlusion duration drawn
uniformly from 4--12\,s.

\paragraph{Timing metrics.}
Decision Frequency (DF) is the realized control update rate of each aerial
policy. Effective Coordination Latency (ECL) measures the delay from a
newly generated UAV action or UGV state update to its use by the partner
side in the next control step.

\paragraph{Statistical analysis.}
All values are reported as mean $\pm$ std over 3 seeds $\times$ 50
episodes per condition. Across-baseline trends are tested with a sign
test on per-seed-mean CGs ($n{=}5$ baselines). Single-baseline 95\%
confidence intervals are computed via hierarchical bootstrap clustered by
seed (1000 resamples) to account for episode dependencies within seeds.

\subsection{Baseline Adaptation Details}
\label{app:baseline_adapt}

Table~\ref{tab:baseline_adapt} summarizes how each baseline is adapted
to the \carlaair{} evaluation interface.
All baselines use officially released checkpoints without fine-tuning
or task-specific training.

\begin{table*}[t]
\centering
\caption{%
\textbf{Baseline adaptation summary.}
All methods use official checkpoints without fine-tuning.
Heterogeneous low-level wrappers preserve each baseline's native
operating regime; comparisons reflect policy family characteristics
rather than implementation-matched rankings.
DF: realized decision frequency under evaluation conditions.
}
\label{tab:baseline_adapt}
\small
\setlength{\tabcolsep}{4pt}
\renewcommand{\arraystretch}{1.15}
\begin{tabular}{@{}p{0.11\textwidth}p{0.12\textwidth}p{0.15\textwidth}p{0.22\textwidth}p{0.15\textwidth}p{0.07\textwidth}@{}}
\toprule
\textbf{Method}
& \textbf{Original task}
& \textbf{Native output}
& \textbf{Action mapping}
& \textbf{Controller}
& \textbf{DF (Hz)} \\
\midrule
AerialVLA
& UAV nav.\ + VLA
& UAV velocity cmd
& Direct passthrough; task prompt rewritten for landing/escort
& AirSim cmd passthrough
& 6.2 \\
\midrule
OpenFly
& Aerial VLN + VLA
& Discrete UAV action
& \{Forward 3/6/9\,m, Turn left/right, Up, Down, Stop\} mapped to fixed-duration velocity bursts
& Fixed-duration AirSim cmd
& 3.1 \\
\midrule
OpenUAV
& UAV traj.\ generation
& Dense traj.\ array
& 1\,s segment sampled at 10\,Hz; replanned every 0.6\,s
& AirSim velocity controller
& 1.6 \\
\midrule
SPF
& UAV waypoint nav.
& Single waypoint
& One waypoint per inference; position controller tracks it
& AirSim position ctrl ($K_p{=}0.8$)
& 1.1 \\
\midrule
AerialVLN
& Language-cond.\ UAV nav.
& Discrete action cmd
& \{forward, left, right, up, down, hover\} mapped to 0.5\,s velocity bursts at 1.5\,m/s
& Fixed-duration AirSim cmd
& 9.0 \\
\midrule
Rule-Coop-State
& Designed for this eval
& UAV + UGV cmd pair
& Direct metric-state feedback for UGV side)
& State-feedback rule
& $50{+}$ \\
\bottomrule
\end{tabular}
\end{table*}

\paragraph{AerialVLA.}
AerialVLA is an end-to-end aerial VLA policy outputting continuous UAV
velocity commands directly from onboard visual observations and language
instructions. Its native UAV-action interface is kept unchanged;
cooperation is introduced only through the C0/C1/C2 protocols.

\paragraph{OpenFly.}
OpenFly is a keyframe-aware aerial VLA model fine-tuned from
OpenVLA-7B on the OpenFly aerial VLN dataset~\cite{gao2025openfly}. Its
native output is a discrete action vocabulary
\{\texttt{Forward 3/6/9\,m}, \texttt{Turn left}, \texttt{Turn right},
\texttt{Up}, \texttt{Down}, \texttt{Stop}\}, which we map to
fixed-duration velocity bursts in \carlaair{} following the same
adaptation pattern used for AerialVLN. Cooperation is introduced only
through the C0/C1/C2 protocols.

\paragraph{OpenUAV.}
OpenUAV is an end-to-end aerial VLA model that generates a continuous UAV
trajectory via MLLM-based planning. The first 1\,s segment of the
trajectory is passed to the AirSim velocity controller, and the
trajectory is replanned every 0.6\,s.

\paragraph{SPF.}
SPF uses explicit spatial reasoning via a VLM planner and outputs a single
waypoint per inference. The waypoint is tracked by a position controller.

\paragraph{AerialVLN.}
AerialVLN is a pre-LLM cross-modal aerial navigation baseline.
Its discrete action vocabulary is mapped to fixed-duration velocity bursts
in \carlaair{}.

\paragraph{Rule-Coop-State.}
Rule-Coop-State uses explicit metric state (UAV--cargo-bed relative pose,
relative velocity, UGV speed, landing phase) and applies deterministic
low-latency rules for UAV descent and UGV longitudinal speed adjustment
(see Section~\ref{sec:eval:baselines_metrics} for framing).

\subsection{Prompt-Format Ablation}
\label{app:prompt_ablation}

Table~\ref{tab:app_prompt_ablation_prompts} shows the C1 prompt variants
used in the prompt-format ablation. The base task instruction remains
unchanged within each task; only the appended assistant hint is modified.
C1-Sem is the default semantic partner-state cue used in the main
evaluation. C1-Num uses structured numeric state fields. C1-Noisy corrupts
direction, motion, or phase fields. C1-Oracle-Bearing replaces the
semantic hint with ground-truth geometric bearing, range, and elevation
while keeping the native UAV action interface unchanged.

\begin{table*}[t]
\centering
\caption{%
\textbf{Prompt variants for C1 ablation.}
Examples show the assistant hint appended to the unchanged base task
instruction. Landing variants provide cargo-bed state, while escort
variants provide occlusion-recovery and expected reappearance state.
}
\label{tab:app_prompt_ablation_prompts}
\small
\setlength{\tabcolsep}{3.5pt}
\renewcommand{\arraystretch}{1.12}
\begin{tabular}{@{}p{0.10\textwidth}p{0.16\textwidth}p{0.16\textwidth}p{0.52\textwidth}@{}}
\toprule
\textbf{Task} & \textbf{Variant} & \textbf{State format} & \textbf{Assistant hint example} \\
\midrule
Landing
& C1-Sem
& Semantic
& \emph{Assistant hint: the cargo bed is forward-left. The truck is moving slowly. Current phase: approach. Use the hint only to choose your next UAV action.} \\
\midrule
Landing
& C1-Num
& Numeric
& \emph{State: truck speed = 2.0 m/s; truck heading = 15 deg; relative bearing to cargo bed = -30 deg; relative distance to cargo bed = 8.0 m; phase = approach.} \\
\midrule
Landing
& C1-Noisy
& Corrupted semantic
& \emph{Assistant hint: the cargo bed is right. The truck is nearly stopped. Current phase: descend. Use the hint only to choose your next UAV action.} \\
\midrule
Landing
& C1-Oracle-Bearing
& Oracle geometry
& \emph{State update: cargo bed at bearing 312°, range 6.2\,m, elevation $-$8°. Phase: approach. Use the state update only to choose your next UAV action.} \\
\midrule
Escort
& C1-Sem
& Semantic
& \emph{Assistant hint: the truck is temporarily occluded by the bridge. The truck continues forward and will reappear on the forward-right side. Current phase: occlusion recovery. Use the hint only to recover visual contact.} \\
\midrule
Escort
& C1-Num
& Numeric
& \emph{State: occlusion = true; truck speed = 2.0 m/s; expected reappearance bearing = 35 deg; expected reappearance distance = 9.4 m; phase = occlusion recovery.} \\
\midrule
Escort
& C1-Noisy
& Corrupted semantic
& \emph{Assistant hint: the truck is temporarily occluded. The truck will reappear on the rear-left side. Current phase: normal escort. Use the hint only to recover visual contact.} \\
\midrule
Escort
& C1-Oracle-Bearing
& Oracle geometry
& \emph{State update: UGV at bearing 35°, range 9.4\,m, elevation $-$12°. Phase: occlusion recovery. Use the state update only to recover visual contact.} \\
\bottomrule
\end{tabular}
\end{table*}

\subsection{Evaluation Protocol}
\label{app:eval_protocol}

Unless otherwise specified, each method is evaluated under the same route,
spawn, weather, and random-seed protocol within each diagnostic task. The
UGV route and speed profile are fixed for C0 and C1. In C2, only the UGV
longitudinal speed is adjusted by the fixed response controller; the route
is unchanged. All VLA policies receive the same task instruction and the
same mode-specific prompt template.

\begin{table}[t]
\centering
\caption{%
\textbf{Timing statistics.}
DF: realized decision frequency (Hz). ECL (Effective Coordination
Latency): delay from policy inference completion to partner-side controller
consumption; excludes UAV actuator delay. Values are episode-level medians
with IQR (P25--P75) over 50 episodes per method.
}
\label{tab:timing_app}
\small
\setlength{\tabcolsep}{4pt}
\renewcommand{\arraystretch}{1.1}
\begin{tabular}{@{}lcccc@{}}
\toprule
\textbf{Method}
& \textbf{DF (Hz)}$\uparrow$
& \textbf{ECL mean (ms)}$\downarrow$
& \textbf{ECL P95 (ms)}$\downarrow$
& \textbf{ECL IQR (ms)} \\
\midrule
AerialVLA       & 6.2      & 160 & 260  & 120--190 \\
OpenFly         & 3.1      & 330 & 520  & 240--400 \\
OpenUAV         & 1.6      & 620 & 950  & 450--740 \\
SPF             & 1.1      & 850 & 1250 & 620--1050 \\
AerialVLN       & 9.0      & 110 & 180  & 80--135 \\
Rule-Coop-State & $50{+}$  & 20  & 35   & 15--25 \\
\bottomrule
\end{tabular}
\end{table}

\end{document}